\documentclass[11pt, letterpaper]{article}

\usepackage[left=1in, right=1in, top=1in, bottom=1in]{geometry}
\usepackage[utf8]{inputenc}
\usepackage[T1]{fontenc}
\usepackage{bm}
\usepackage{type1cm}
\usepackage{lettrine}
\usepackage{amsmath,amssymb,amsthm}
\usepackage{moreverb}
\usepackage{mathtools}
\usepackage{amsmath}
\usepackage{amssymb}
\usepackage{algorithmic}
\usepackage{graphics}
\usepackage{graphicx}
\usepackage{subfigure}
\usepackage{caption}
\usepackage{extarrows}
\usepackage{color}
\usepackage{framed}
\usepackage{wrapfig}
\usepackage{bm}
\usepackage{mathrsfs}
\usepackage{mathabx}
\usepackage{multirow}
\usepackage{longtable}
\usepackage{hyperref}
\usepackage{paralist}
\usepackage{indentfirst}
\usepackage{relsize}
\usepackage{extarrows}
\usepackage{upgreek}
\usepackage{bm}
\usepackage{mwe}
\usepackage[dvipsnames,table]{xcolor}
\usepackage{booktabs}
\usepackage{authblk}
\usepackage{lettrine}
\usepackage{type1cm}
\usepackage{threeparttable}
\usepackage[sort&compress,numbers]{natbib}
\usepackage[figurename=Figure]{caption}
\usepackage[normalem]{ulem}
\usepackage{adjustbox}
\usepackage{ulem}
\usepackage{xcolor}

\graphicspath{ {./Figure/} }
\usepackage[font=footnotesize,labelfont=bf]{caption}

\newcommand{\graph}[1]{\mathcal{}}
\providecommand{\keywords}[1]{\textbf{\textit{Keywords: }} #1}

\hypersetup{
bookmarks=true,
bookmarksopen=true,
bookmarksnumbered=true,
unicode=false,
pdftoolbar=true,
pdfmenubar=true,
pdffitwindow=false,
pdfstartview={FitH},
pdftitle={My title},
pdfauthor={Author},
pdfsubject={Subject},
pdfcreator={Creator},
pdfproducer={Producer},
pdfkeywords={keywords},
pdfnewwindow=true,
colorlinks=true,
linkcolor=blue,
citecolor=blue,
filecolor=blue,
urlcolor=blue
}

\begin{document}
\title{\textbf{Perovskite‑R1: A Domain‑Specialized LLM for Intelligent Discovery of Precursor Additives and Experimental Design}}
\author[1,$\dag$]{Xin-De Wang}
\author[2,$\dag$] {Zhi-Rui Chen}
\author[1]{Peng-Jie Guo}
\author[1,*]{Ze-Feng Gao}
\author[2,*] {Cheng Mu}
\author[1,*]{Zhong-Yi Lu}

\affil[1]{\small School of Physics, Renmin University of China, Beijing, China}
\affil[2]{\small School of Chemistry and Life Resource, Renmin University of China, Beijing, China \vspace{18pt}}
\affil[$\dag$]{Equally contributed\vspace{6pt}}
\affil[*]{Corresponding authors\vspace{12pt}}

\date{}

\maketitle

\normalsize

\vspace{-28pt} 
\begin{abstract}
\small
Perovskite solar cells (PSCs) have rapidly emerged as a leading contender in next-generation photovoltaic technologies, owing to their exceptional power conversion efficiencies and advantageous material properties. Despite these advances, challenges such as long-term stability, environmental sustainability, and scalable manufacturing continue to hinder their commercialization. Precursor additive engineering has shown promise in addressing these issues by enhancing both the performance and durability of PSCs. However, the explosive growth of scientific literature and the complex interplay of materials, processes, and device architectures make it increasingly difficult for researchers to efficiently access, organize, and utilize domain knowledge in this rapidly evolving field. To address this gap, we introduce Perovskite-R1, a specialized large language model (LLM) with advanced reasoning capabilities tailored for the discovery and design of PSC precursor additives. By systematically mining and curating 1,232 high-quality scientific publications and integrating a comprehensive library of 33,269 candidate materials, we constructed a domain-specific instruction-tuning dataset using automated question–answer generation and chain-of-thought reasoning. Fine-tuning the QwQ-32B model on this dataset resulted in Perovskite-R1, which can intelligently synthesize literature insights and generate innovative and practical solutions for defect passivation and the selection of precursor additives. Experimental validation of several model-proposed strategies confirms their effectiveness in improving material stability and performance. Our work demonstrates the potential of domain-adapted LLMs in accelerating materials discovery and provides a closed-loop framework for intelligent, data-driven advancements in perovskite photovoltaic research.
\end{abstract}

\vspace{12pt} 

\section*{Introduction}\label{sec1}
In recent years, perovskite solar cells (PSCs) have garnered significant attention from both academia and industry due to their tremendous potential in sustainable energy applications. Since their introduction in 2009, the power conversion efficiency (PCE) of PSCs has rapidly increased from 3.8\% to 26.95\%~\cite{kojima2009organometal,NRELCellEfficiency2025}, highlighting their promising prospects as next-generation photovoltaic technology. The remarkable efficiency improvement of PSCs is mainly attributed to the unique advantages of perovskite materials, such as highly ordered crystal structures, excellent light absorption, tunable band gaps, and cost-effective solution-based fabrication~\cite{bian2025removal,tian2025divalent}. These features not only enable high performance but also facilitate the scalability of PSCs for large-scale applications.

However, the commercialization of PSCs still faces significant challenges, including long-term operational stability, environmental sustainability, and scalable manufacturing~\cite{hassan2022recent,afroz2025perovskite}. Issues such as sensitivity to moisture and heat, material degradation, and process compatibility remain key bottlenecks limiting practical deployment. To address these challenges, precursor additive engineering has emerged as an effective strategy to enhance both the stability and efficiency of PSCs. By introducing suitable additives into the precursor solution, researchers can optimize crystallization, passivate defects, and improve film quality, thereby significantly improving device performance and durability. This approach offers strong support for the further industrialization of PSC technology.

Traditionally, progress in PSCs research has relied on researchers conducting comprehensive literature reviews and then applying intuition and experience-based judgment to analyze findings, followed by experimental validation. However, the rapid advancement of PSCs in recent years has led to an exponential increase in the number of related publications, making it increasingly challenging for researchers to efficiently access and utilize the ever-growing body of knowledge in this field. This challenge is particularly acute given the complex interplay among material composition, fabrication processes, and device architecture that characterize PSCs research. On the other hand, existing artificial intelligence systems~\cite{yang2024discovering,wang2023feature,abdellah2024machine} in materials science typically focus on specific prediction tasks or general scientific knowledge~\cite{liu2025perovskite,kim2025explainable}, lacking the specialized capability to address the unique characteristics of PSC research~\cite{han2025perovskite}. This gap highlights the urgent need for an integrated system that can systematically organize domain knowledge and provide intelligent assistance to researchers.

Recently, large language models (LLMs) have developed rapidly, with numerous high-performance models emerging~\cite{hurst2024gpt,team2023gemini,touvron2023llama,liu2024deepseek,bai2023qwen}. Advances in LLMs are primarily driven by increases in model size, the scale of pre-training data, and enhanced reasoning capabilities. Recently, much attention has focused on improving reasoning abilities, which not only boost accuracy but also enhance interpretability and controllability. For example, the introduction of Chain-of-Thought (CoT) prompting in 2022 led to significant improvements in mathematical reasoning tasks for large models~\cite{wei2022chain}, laying the groundwork for further developments in the field. Moreover, integrating CoT methods with reinforcement learning has further advanced model autonomy, enabling superior performance in tasks such as mathematics, programming, and scientific reasoning~\cite{jaech2024openai,guo2025deepseek,yang2025qwen3}. At the same time, applying LLMs to specialized domains has become an important trend. However, due to complex terminologies and extensive knowledge structures in these fields, general-purpose models often fall short. Techniques such as domain-specific fine-tuning and structured knowledge injection have facilitated the emergence of specialized LLMs. For instance, dedicated LLMs have been developed and applied in areas such as biomedicine, materials science, education, and finance~\cite{luo2023biomedgpt,kim2025medbiolm,chen2024pharmagpt,antunes2024crystal,gruver2024fine,sriram2024flowllm,mishra2024foundational,qu2024coursegpt,jiang2024beyond,yang2023investlm,yang2024financial,liu2025fin}.

In this work, we present Perovskite-R1, an LLM with advanced reasoning capabilities designed specifically for the discovery and design of PSC precursor additives. Perovskite-R1 aims to provide intelligent and systematic support for material design and screening in the field of perovskite photovoltaics. Leveraging the capabilities of Perovskite-R1, we systematically identify a set of potential defect compensation strategies for perovskite precursor additives. Several of these strategies are further validated through experimental investigations, which demonstrated that the solutions proposed by Perovskite-R1 are both rational and reliable, leading to improvements in material performance and stability. Specifically, we collect 1,232 scientific publications related to perovskite precursor design as well as a candidate library containing 33,269 materials. Utilizing the OpenAI o1 model\footnote{The full version of OpenAI o1 model was released on December 5, 2024.}\cite{jaech2024openai}, we transform the content of these papers into an instruction-tuning dataset in the form of question–answer pairs, thereby enriching the model’s domain-specific knowledge and practical examples. Building on this high-quality instruction dataset, we fine-tune the QwQ-32B pre-trained model\footnote{The QwQ-32B model was first released on March 6, 2025.}\cite{qwq32b} to develop Perovskite-R1. Thanks to this comprehensive dataset and instruction tuning, Perovskite-R1 can effectively integrate existing knowledge on perovskite precursors with the latest research advances and experimental data, generating innovative and practical design solutions. 

To experimentally validate the predictive power of our model, we conducted a systematic comparison between additives recommended by Perovskite-R1~(3, 5-difluoropyridine-2-carboxylic acid (AI-DFCA) and 5-hydroxy-2-methylbenzoic acid (AI-HMBA)) and those chosen manually by researchers~(gallic acid (Manual-GA) and caffeic acid (Manual-CA)). All additives were incorporated at equal concentrations into Cs$_{0.05}$MA$_{0.1}$FA$_{0.85}$PbI$_{3}$ perovskite devices under identical fabrication conditions. The results revealed that model-identified additives significantly improved device performance, while manually selected additives led to inferior outcomes. This highlights the advantage of data-driven screening over traditional, experience-based approaches in complex materials discovery, opening new avenues for intelligent material discovery in the PSC field.

\section*{Related work}\label{sec-related-work}

\paragraph{AI technology for materials science} Research in materials science is characterized by a high degree of diversity and complexity. Consequently, the development of computational tools for material properties and the discovery of new materials have been a primary focus of research efforts. In light of the recent advancements in AI, it becomes imperative to explore the potential of AI technology in the domain of materials science~\cite{han2024ai,li2025materials,cheng2025ai}.

A notable example of work on predicting material properties using AI is ALIGNN~\cite{choudhary2021atomistic}. By incorporating bond angle information into graph neural networks, ALIGNN significantly improves the accuracy of material property predictions. Due to its excellent performance and ease of training, ALIGNN is still used in many studies today~\cite{alkabakibi2025graph,sibi2024advancing,ojih2024graph,gurunathan2023rapid}. In addition to ALIGNN, many other efforts have been made to predict material properties~\cite{du2024ctgnn,wen2024equivariant,qayyum2023explainable}. Each method has its own strengths and weaknesses, and specialises in a particular area. To this day, material prediction methods applied to different fields are still being developed.

Another important application of AI in materials science is reverse design. This area has seen a lot of excellent work. An earlier work is Crystal Diffusion Variational Autoencoder~(CDVAE), which uses a diffusion process to generate crystal structures with SE(3) and periodic invariance~\cite{xie2021crystal}. This process optimizes specified properties, such as final energy states. Another excellent work is DiffCSP~\cite{jiao2023crystal}. It provides greater stability and accuracy in structural prediction by combining diffusion lattices and fractional coordinates, as well as introducing improvements to degeneracy, such as E(3) equivariance. GNoME is a significant work that combines graph neural networks~(GNNs) and density functional theory to successfully predict more than 380,000 novel stabilized materials, thereby significantly expanding the number of known stabilized materials~\cite{merchant2023scaling}.

\paragraph{LLM for materials science}
In consideration of the recent advancements in LLM, there have been several attempts to apply this technique to the materials domain. LLMs exhibit greater flexibility and maneuverability in comparison to diffusion models and GNNs, suggesting significant potential in the field of materials research~\cite{han2024ai,cheng2025ai}.

A conventional approach involves training a materials-specific LLM from the scratch with the assistance of a materials database, as exemplified by CrystaLLM~\cite{antunes2024crystal}. CrystaLLM is an autoregressive LLM that has been trained on millions of crystal structure (CIF) files to generate plausible inorganic crystal structures from textual prompts. Moreover, there are other studies that adopt a similar approach. These studies borrow fine-tuning techniques~\cite{gruver2024fine} or combine them with flow-matching models~\cite{sriram2024flowllm}. These techniques reduce the computational power requirement during training, enhance the model, and improve its capabilities.

In addition to the aforementioned works, numerous recent studies have adopted techniques and methodologies from the LLM field, thereby facilitating novel advancements in materials science. MatterChat employs a multimodal LLM that integrates the atomic structure of a material with the text, thereby facilitating property prediction and experimental process reasoning~\cite{tang2025matterchat}. HoneyComb is a materials science-specific LLM intelligence system that integrates tool invocation, structured knowledge base, and retrieval modules to improve the accuracy and adaptability of complex tasks, such as computational tasks or documentation~\cite{zhang2024honeycomb}. LLaMP is an integrated platform that combines a material project database and high-throughput simulation~\cite{chiang2024llamp}. It supports crystal structure editing, molecular dynamics simulation calling, and high-precision property prediction. LLaMP is based on the Retrieval-augmented Generation~(RAG) and ReAct agents, which significantly reduces the phantom problem. 
\section*{Results}\label{sec-results}
\begin{figure}[t]
    \centering
    \includegraphics[width=\linewidth]{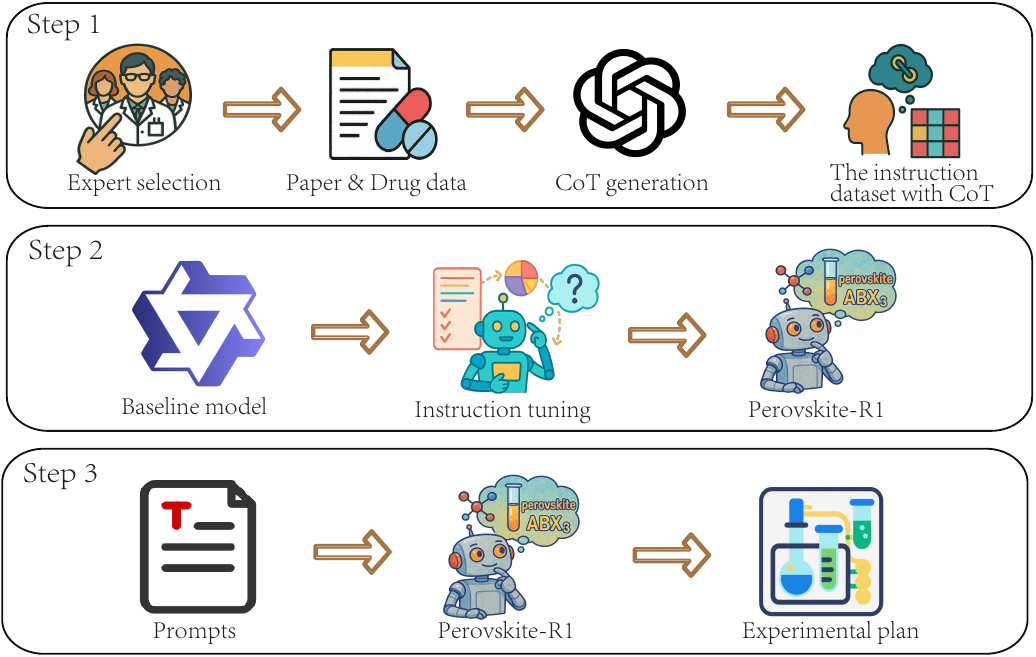}
    \caption{\textbf{The detail of the construction of the proposed Perovskite-R1.} In step 1, we select the proper papers and drug data, and then use an LLM to construct a dataset containing CoT. In step 2, we use scientific journal articles and drug data to conduct instruction tuning the baseline model. Finally, in step 3, we design prompts to obtain the desired experimental protocols. }
    \label{fig:main-fig}
\end{figure}

We train a domain-specific LLM, Perovskite-R1, focusing on defect passivation in the field of perovskite materials, with the goal of enhancing its scientific reasoning and generation capabilities for tasks such as the selection of precursor additives and experimental design. To achieve this, we develop a comprehensive framework encompassing high-quality dataset construction, supervised fine-tuning with CoT annotations, structured prompt engineering, and systematic performance evaluation. As illustrated in Figure~\ref{fig:main-fig}, the framework integrates curated perovskite literature and a drug-like compound library, ensuring both domain coverage and improved interpretability and generalization in complex materials science tasks. We further construct a benchmark dataset and compare Perovskite-R1 against several leading LLMs, demonstrating its superior performance in addressing perovskite-related challenges. Finally, we showcase the application of Perovskite-R1 in discovering high-performance defect-passivating precursor additives, highlighting its potential as an intelligent engine for next-generation materials discovery. The following subsections provide a detailed account of these key components and results.

\subsection*{Framework of Perovskite-R1}

The overall construction process of Perovskite-R1 can be divided into four key stages. First, a high-quality instruction fine-tuning dataset is built through systematic literature mining. Second, based on this dataset, the LLM is fine-tuned using the LoRA strategy, resulting in the Perovskite-R1 model with specialized expertise in perovskite materials. Third, structured prompts are designed to guide the model in specific tasks such as experimental design and material screening. Finally, the model-generated recommendations are implemented as practical experimental protocols, which are then synthesized and validated in the laboratory, thus establishing a closed-loop system from language model generation to physical experimentation.

The foundation of Perovskite-R1 lies in the construction of a high-quality instruction fine-tuning dataset specifically oriented toward the task of defect passivation in perovskite precursors. To achieve this, we meticulously curate a selection of 1,232 peer-reviewed research articles from top scientific publishers, including the American Chemical Society (ACS), Elsevier, the Royal Society of Chemistry (RSC), Wiley, and Springer Nature. Our selection criteria prioritize studies that provided detailed descriptions of precursor additive synthesis pathways, material characterization methods, and performance optimization strategies, with a particular emphasis on works demonstrating significant achievements in device efficiency or stability and publications featured in high-impact journals. To ensure comprehensive coverage of both foundational and cutting-edge knowledge, we trace the evolution of perovskite precursor additive strategies from their early development to the latest breakthroughs, thus capturing the full temporal spectrum of the field. Additionally, to enhance the model’s generalization across chemical space, we incorporate a large-scale drug-like molecular library containing 33,269 distinct compounds. This resource not only furnishes the model with a diverse array of candidate additive structures, but also enables it to learn the intricate relationships between molecular structures and material properties, thereby strengthening its chemical reasoning and generative capabilities.

During the data preprocessing stage, we utilize the pdfminer tool to convert the selected PDF documents into structured plain text. To balance contextual coherence and computational efficiency, each paper is segmented into text chunks of up to 2,500 tokens, with a 20\% overlap between adjacent segments to maintain semantic consistency across paragraphs. This strategy not only enhances the logical integrity of the text sequences but also significantly improves the coherence and quality of CoT reasoning paths in subsequent question–answer generation tasks. Next, we employ the OpenAI o1 API to automatically generate high-quality question–answer pairs from these text segments. To further strengthen the model’s scientific reasoning abilities, we introduce a customized CoT template that guides the model to produce responses containing explicit reasoning steps along with final conclusions. As a result, the original corpus is transformed into an instruction-tuning dataset with deep semantic structure and scientific logic, endowing the model with advanced capabilities in experimental design and mechanistic understanding of materials.

For the model training stage, we select the powerful QwQ-32B model~\cite{qwq32b} as the base language model and perform efficient parameter fine-tuning using the LoRA technique on the aforementioned CoT-enhanced dataset. By introducing low-rank matrices for targeted adjustments to the original model parameters, this approach significantly reduces training costs while effectively embedding specialized knowledge from the perovskite domain. As a result, the model internalizes and transfers domain knowledge relevant to specific materials science tasks. The fully trained Perovskite-R1 model thus retains robust general language modeling capabilities while acquiring expert-level proficiency in tasks such as materials structure design, defect regulation, and synthesis pathway recommendation.

To enable controlled application of the model in real-world scientific research tasks, we develop a structured prompt system tailored to the research workflow. Each prompt is divided into three components: task definition (e.g., Select suitable additives for defect passivation in FAPbI$_{3}$), scientific and application-oriented criteria (e.g., Prioritize heterocyclic compounds that can enhance device stability), and output format specification (e.g., Please provide the molecular structure and its anticipated mechanism of action). This prompt system supports both single-turn and iterative multi-turn interactions, allowing for interactive scientific dialogue with the model. Ultimately, under the guidance of materials chemistry experts, we design several experimental protocols based on the recommendations generated by Perovskite-R1 and carried out corresponding synthesis and validation in the laboratory. The results demonstrate that some of the novel additives suggested by the model exhibited excellent performance in defect passivation and stability enhancement, confirming their ability to provide innovative experimental insights for practical research applications.

\subsection*{Maintenance of reasoning ability}

For the molecular library data, compounds are grouped into sets to ensure diversity and manageability. The o1 model is then tasked with interpreting the properties of each compound within a group using CoT-style reasoning, and subsequently identifying which molecules could serve as promising precursor additives. The resulting question–answer data from both the literature analysis and molecular library are merged to form a comprehensive fine-tuning dataset for QwQ-32B. Through this workflow, we derive the specialized Perovskite-R1 model.

\begin{figure}
    \centering
    \includegraphics[width=\linewidth]{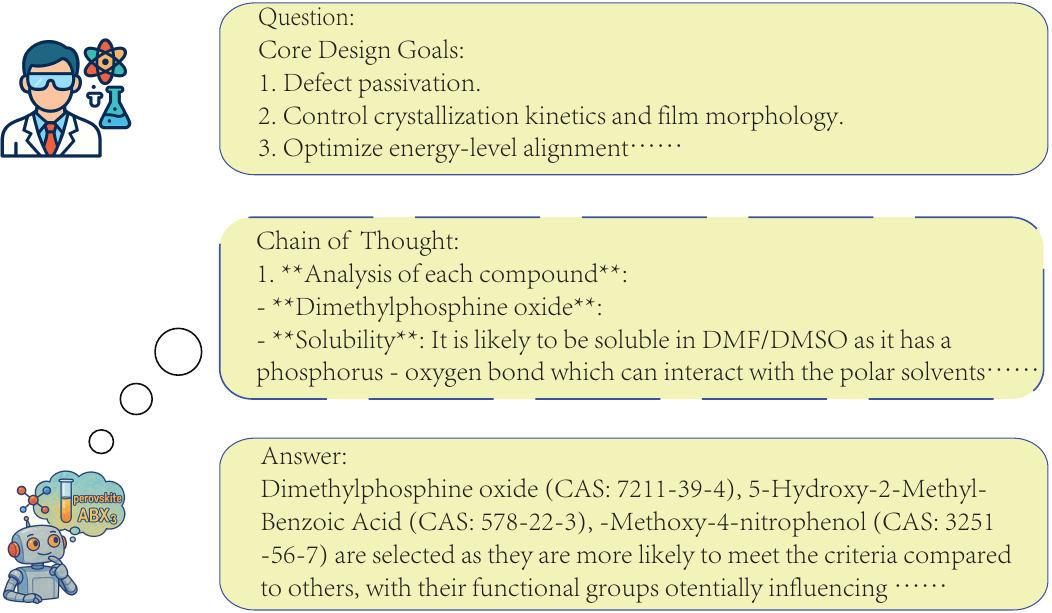}
    \caption{\textbf{An example of dialogue.} The user provides the prompt, then Perovskite-R1 first gives its thought process and following its final result. }
    \label{fig:fig-2}
\end{figure}

This carefully orchestrated process ensures that the model’s advanced reasoning abilities are not only preserved but also enhanced in the context of the discovery of perovskite precursor additives. As illustrated in Figure~\ref{fig:fig-2}, Perovskite-R1 is thus capable of synthesizing complex scientific knowledge and generating actionable, high-quality recommendations for experimental research.

\subsection*{Construction of prompts}
The effectiveness of LLM outputs is highly dependent on the quality and design of the prompts provided during inference. Recognizing this crucial factor, we adopt a carefully crafted prompt strategy in our work. Our prompts are methodically structured to ensure clarity, precision, and contextual relevance, while also encompassing a broad spectrum of domain-specific expertise. By systematically incorporating task definitions, scientific criteria, and explicit output requirements, our approach not only guides the LLM toward generating accurate and insightful responses but also enhances its ability to address complex, specialized research questions with greater reliability and depth. The prompt we design consists of three clearly delineated sections:
\begin{enumerate}
\item \textbf{Task Definition.} In the first section, we explicitly state the overarching goal: guiding the model to recognize and select appropriate chemical precursor additives for perovskite synthesis. For instance, the core mission of Perovskite-R1 can be conveyed through the formulation of specific objectives or requirements to be accomplished within this section.
\item \textbf{Objective Criteria.} The second section lays out the scientific and application‑driven requirements. In this section, the focus is on distinguishing Perovskite-R1 from a standpoint that determines the suitability of additives or the specific requirements for the desired additives. This standpoint encompasses the reaction process that the additives are expected to fulfil and the properties they are required to exhibit, to the greatest extent possible.
\item \textbf{Expected Output Specification.} In the final section, we detail exactly what the model’s response should include. As a case in point, the format and amount of perovskite-R1 output content can be specified, along with an analysis of the relevant choices.

\end{enumerate}

In addition to the three primary points of the appeal, it is possible to add some additional requirements that may be valid. These could include the requirement for Perovskite-R1 to carefully analyse each additive, or the requirement for a suitable additive to not be made without careful consideration.

Moreover, given the extensive collection of drug libraries, it is imperative to ensure that perovskite-R1 has thoroughly evaluated all available drugs. To this purpose, a systematic approach involving traversal, screening, and subsequent iteration is adopted. The procedure entails the initial classification of all drugs into equal numbers of distinct groups. Thereafter, up to three additives from each group are selected based on their suitability. These additives are then subjected to a subsequent screening step, wherein they are evaluated within their respective groups. This process is repeated until a satisfactory number of additives are identified. This strategy ensures that our drug library is fully utilised and demonstrates the superiority of LLM in screening judgment, as this task is difficult to accomplish in a short period of time by human labour, whereas Perovskite-R1 may be able to do it in a few hours or even minutes.

\subsection*{Experiment of Perovskite-R1}

To rigorously evaluate the effectiveness of our model and enable fair comparison with other state-of-the-art language models, we constructed a comprehensive benchmark dataset focused on perovskite research. This dataset comprises a wide range of carefully curated question–answer pairs that span various domains of perovskite science. Each question was independently assessed by domain experts and categorized into three levels of difficulty—easy, medium, and hard—based on the complexity of the scientific concepts and the reasoning required.
 
We subsequently benchmark Perovskite-R1 against several prominent language models, including DeepSeek-R1, Gemini-2.5-Flash-Thinking, and others, using this dataset as a standardized evaluation platform. As summarized in Table~\ref{tab:benchmark}, Perovskite-R1 consistently outperforms the competing models across all difficulty categories, with particularly excelling in challenging, domain-specific tasks. Notably,  Perovskite-R1 has achieved a 10\% enhancement in accuracy over the original QwQ-32B. These results provide compelling evidence for the superior capability and domain adaptation of Perovskite-R1 in addressing perovskite-related scientific problems, thereby substantiating both the efficacy and practical relevance of our approach.

\begin{table}[htbp]
  \centering
  \begin{threeparttable}
    \begin{tabular}{lccc}  
      \toprule
      \multirow{2}{*}{\bfseries Model} &
        \multicolumn{3}{c}{\bfseries Benchmark} \\  
      & \bfseries Easy & \bfseries Medium & \bfseries Hard \\  
      \midrule
      QwQ-32B & 76.27 & 76.84 & 74.75 \\
      Doubao-seed-1.6-thinking & 71.30 & 71.46 & 70.77 \\
      DeepSeek-R1 & 76.42 & 74.88 & 73.97 \\     
      Gemini-2.5-flash-thinking & 78.71 & 77.52 & 76.05 \\
      Perovskite-R1 & \textbf{86.92} & \textbf{85.06} & \textbf{84.63} \\     
      \bottomrule
    \end{tabular}
  \end{threeparttable}
  \caption{The performance of the Perovskite-R1 on benchmark dataset. 
  }
  \label{tab:benchmark}
\end{table}

\subsection*{Discovery of suitable precursor additives}

A primary objective in the development of Perovskite-R1 is to enable the identification of high-quality precursor additives for the synthesis of advanced perovskite materials. This section demonstrates the practical application of Perovskite-R1 in guiding the discovery and screening for precursor additives.

Utilizing carefully designed prompts that integrate multiple evaluation criteria, we conduct a comprehensive screening process with Perovskite-R1. The model efficiently generated a targeted shortlist of small-molecule additives that satisfy the specified requirements and are predicted to enhance both device efficiency and operational stability. These high-potential candidates are currently undergoing systematic experimental validation. This involves formulating precursor solutions, characterizing film morphology and optoelectronic properties, and fabricating complete solar cell devices to rigorously assess the actual impact of each additive on PCE, film quality, and long-term device performance. This closed-loop framework highlights Perovskite-R1's effectiveness in accelerating the discovery and evaluation of promising perovskite precursor additives.

\begin{figure}
    \centering
    \includegraphics[width=1\linewidth]{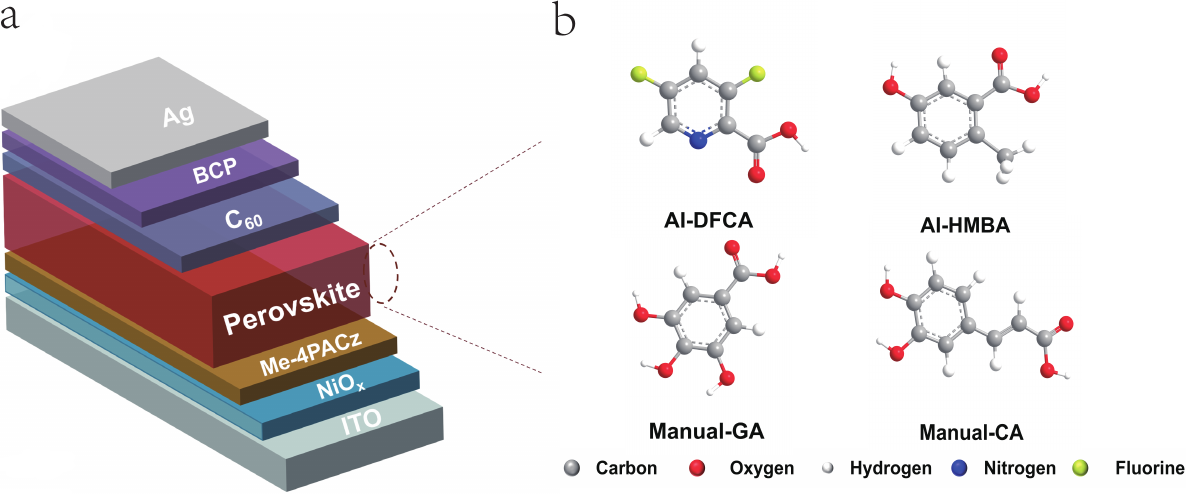}
    \caption{\textbf{Device architecture and the selected moleculars.} a. Schematic of the PSC structure. b. The four precursor additives selected for analysis are DFCA and HMBA, which were chosen by Perovskite-R1, and GA and CA, which were chosen manually.}
    \label{fig:fig-3}
\end{figure}

To empirically validate the model's predictions, a comparative experiment was conducted. Two types of additives were selected for systematic comparison. The first includes 3, 5-difluoropyridine-2-carboxylic acid (AI-DFCA) and 5-hydroxy-2-methylbenzoic acid (AI-HMBA), which were autonomously recommended by Perovskite-R1. The second comprises gallic acid (Manual-GA) and caffeic acid (Manual-CA), manually selected by researchers from the same molecular library based on domain knowledge, existing literature reports and experience. The chemical structure of all four additives is shown in Figure~\ref{fig:fig-3}.

To ensure experiment accuracy, all four additives were introduced into the precursor solution of the perovskite component Cs$_{0.05}$MA$_{0.1}$FA$_{0.85}$PbI$_{3}$ at the same concentration (0.1 mg/mL), and the photovoltaic devices were fabricated using procedures, as shown in Figure~\ref{fig:fig-3}. PCE measurements were conducted under standard conditions (AM 1.5G, 100 mW/cm$^{2}$).

\begin{figure}[t]
    \centering
    \includegraphics[width=1\linewidth]{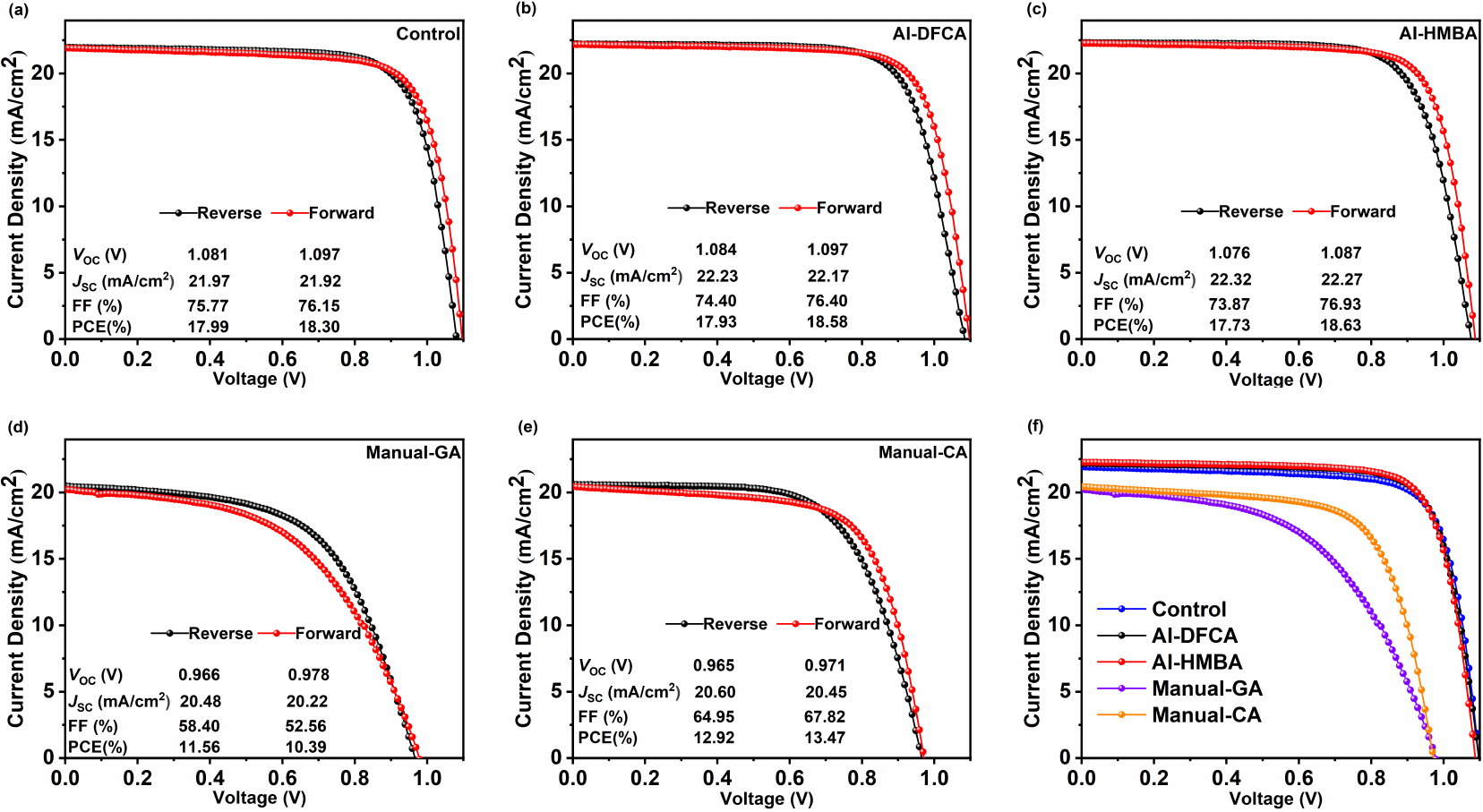}
    \caption{\textbf{Current density-voltage (\textit{J–V}) characteristic curves (forward and reverse scans) of PSCs.} As illustrated in (a), the control group is represented, while (b) through (e) present the experimental outcomes for the four selected molecules. And (f) offers a comprehensive summary of the experimental results. Test conditions: scan rate of 10mV/10ms, AM1.5G illumination~(100 mW/cm$^{2}$). The open-circuit voltage~($V_{OC}$), short-circuit current density~($J_{SC}$), fill factor~(FF), and PCE for both forward and reverse scans are labeled in the figure.}
    \label{fig:fig-4}
\end{figure}

Figure~\ref{fig:fig-4} presents the \textit{J–V} curves of the champion devices with and without the additive treatment. The experimental results in Table~\ref{tab:performance_metrics} show a clear and highly consistent trend. Both DFCA and HMBA, the additives selected by Perovskite-R1, enhance the device performance, achieving PCEs of  18.58\% and 18.67\%, respectively. This result directly validates the model’s ability to identify effective additives from a vast number of candidate molecules. In sharp contrast, the manually selected additives Manual-GA and Manual-CA demonstrate detrimental effects on device performance, reducing the PCE to 11.56\% and 13.47\%, respectively, substantially below the baseline value of 18.30\% observed in the control devices. This contrast underscores the limitations of manual selection, even when guided by expert knowledge. It further emphasizes that cognitive biases and gaps in information coverage can significantly hinder additive screening in complex material systems.

\begin{table}[h]
    \centering
    \begin{tabular}{lccccc}
        \toprule
        Additives & $V_{OC}$ (V) & $J_{SC}$ (mA/cm$^{2}$) & FF (\%) & PCE (\%) \\
        \midrule
        Control   & 1.097 & 21.92 & 76.15 & 18.30 \\
        Manual-GA & 0.966 & 20.48 & 58.40 & 11.56 \\
        Manual-CA & 0.971 & 20.45 & 67.82 & 13.47 \\
        AI-DFCA   & 1.097 & 22.17 & 76.40 & 18.58 \\
        AI-HMBA   & 1.087 & 22.27 & 76.93 & 18.63 \\
        \bottomrule
    \end{tabular}
    \caption{The photovoltaic parameters for PSCs with different additives. 
    }
    \label{tab:performance_metrics}
\end{table}

Importantly, since all four experiments followed identical device fabrication processes and test conditions, the observed differences in efficiency can be directly attributed to the additive selection strategy. This set of data presents a strict opposing pattern: the two additives recommended by Perovskite-R1 both significantly improve the device efficiency, while the two additives selected manually both show inhibitory effects. This direct comparison reinforces that even expert-driven manual selection may fail to predict optimal candidates, whereas Perovskite-R1 has the ability to identify effective additives from a vast number of candidate molecules.

These findings provide strong evidence that Perovskite-R1 developed in this study significantly outperforms the manual screening method that relies on the subjective experience of researchers in the initial screening efficiency of perovskite precursor solution additives. The ability of the model to correctly predict the performance direction of additives demonstrates its high reliability and practical relevance.

The success of Perovskite-R1 stems from its in-depth learning of the structure-activity relationship between molecular features and additive functions, based on insights extracted from 1,232 professional and  33,269 molecular features. This comprehensive training enables the model to break through the limitations of individual cognition and the blindness of the traditional trial-and-error method, significantly improving the screening efficiency and reducing development cost. Overall, this result confirms the huge potential of AI-based models in solving complex material screening problems and accelerating the development of perovskite photovoltaic materials, providing a new approach for breaking through the traditional scientific research model.

\section*{Discussion}\label{sec12}

In this work, we address the inefficiency and reliance on expert intuition in designing defect compensation strategies for PSC precursor additives by introducing Perovskite-R1, a domain-specific LLM with advanced reasoning capabilities. We construct a high-quality instruction-tuning dataset comprising 1,232 scientific publications and 33,269 candidate materials, and fine-tune the QwQ-32B base model to develop Perovskite-R1. Unlike conventional static prediction models, Perovskite-R1 systematically integrates domain knowledge and up-to-date experimental findings to generate innovative and practical material design solutions. Experimental validation confirms the model’s effectiveness, with several proposed strategies significantly enhancing material stability and device performance. This work not only advances the intelligent design of PSC materials but also provides a transferable framework for building high-performing domain-specific language models in other scientific fields.

Although Perovskite-R1 can accurately recognize terminology and provide reasonable answers in single-turn question answering, the depth of its output is limited by the phrasing of the question, often remaining superficial. For example, when addressing complex mechanisms such as defect passivation, the model tends to offer concise summaries without delving into detailed explanations. To address this issue, a preliminary solution is to introduce a multi-turn dialogue mechanism, enabling the model to gradually reason and elaborate through continuous questioning and contextual review, thus achieving more comprehensive and in-depth scientific analysis. In the future, further improvements can be made by incorporating reinforcement learning (such as RLHF) to optimize dialogue guidance strategies and building a multi-turn dialogue knowledge tracking module to enhance the model’s ability to understand consecutive questions. Besides, the control over structured experimental design tasks still has certain limitations. Currently, the model primarily generates experimental protocols through text prompts. However, parameter settings such as molar concentration, additive ratio, and spin‑coating speed still require manual post‑processing, as the model does not precisely model process feasibility or preparation‑condition windows. As an initial solution, a structured prompt template has been developed, consisting of three main components: task definition, target criteria, and output format, which significantly improves the standardization of experimental recommendations. For future development, the introduction of a knowledge graph-based feasibility constraint checking module is planned, along with the integration of physical simulation models to provide parameter space mapping and automatic verification of boundary conditions.

Looking ahead, the development of Perovskite-R1 will focus on several key directions, including expanding the coverage of material systems, enhancing multi-task capabilities, and improving the synthesizability and experimental closed-loop capability of generated solutions. The current model centers on the design of defect passivation strategies for perovskite precursor additives, and has already demonstrated both intelligence and practicality in small-molecule additive screening and experimental validation. In the future, we aim to broaden the application boundaries of the model to encompass a wider range of tasks, such as interface engineering design, solvent system optimization, stability regulation strategies, and the co-design of device architectures. Leveraging Perovskite-R1’s strengths in chemical structure reasoning and condition recommendation, its applications can also be extended to the intelligent development of low-dimensional perovskites, perovskite-silicon tandem solar cells, and stable wide-bandgap material systems.

More importantly, Perovskite-R1 establishes an efficient closed-loop framework that integrates literature mining, knowledge consolidation, solution generation, and experimental validation. This paradigm can be further generalized to other functional material fields, especially in the design of emerging energy storage materials (such as solid-state electrolytes and magnesium-based hydrogen storage materials), catalytic materials (for example, CO$_2$ reduction and electrocatalytic oxygen evolution), and flexible electronic materials (including polymers with strain-responsiveness and self-healing properties). By integrating with automated synthesis platforms and high-throughput experimental systems, Perovskite-R1 can further evolve into a ``research assistant-type intelligent agent'' that actively proposes hypotheses, validates feedback, and self-optimizes, thus propelling materials science from a human-centered, experience-driven exploration model to a model-centered, autonomous innovation paradigm.

\section*{Methods}

\paragraph{Large language model}

For our baseline, we select QwQ-32B, a state-of-the-art causal language model based on the Qwen2.5-32B architecture, featuring approximately 32.5 billion parameters, of which 31 billion are dedicated to non-embedding computations~\cite{qwq32b}. The model is composed of 64 Transformer layers and incorporates a range of advanced architectural features, including Rotary Position Embedding (RoPE) for improved positional encoding, SwiGLU activation functions for enhanced non-linearity, RMSNorm for stable and efficient normalization, and Attention QKV bias for fine-tuning attention distributions.

One of the distinguishing aspects of QwQ-32B is its adoption of grouped-query attention (GQA), which utilizes 40 query heads partitioned into eight shared key–value heads. This design enables the model to achieve efficient and scalable sparse attention, making it capable of processing exceptionally long sequences—up to 131,072 tokens—without compromising computational efficiency. To adapt QwQ-32B for domain-specific tasks, we perform instruction tuning (supervised fine-tuning) using our curated perovskite instruction dataset, thereby endowing the model with specialized knowledge and reasoning abilities tailored to the discovery and design of PSC precursors.

\paragraph{The construction of dataset}
To construct a comprehensive and domain-specific training corpus, we curate a total of 1,232 research articles alongside a library of 33,269 candidate drug compounds relevant to perovskite precursor design. Due to the model’s fixed input token limitations, each research article is systematically segmented into passages of uniform length, yielding approximately 5,000 discrete text fragments for subsequent processing. In parallel, all candidate compounds—systematically identified by both their common names and CAS registry numbers—are classified into seven major therapeutic categories, reflecting the breadth of chemical diversity considered. This organizational strategy result in 4,753 individual data records representing alternative chemical agents. Together, these structured resources form a robust foundation for high-quality instruction tuning, ensuring that the model is exposed to a wide variety of scientific contexts and molecular candidates during training.

\begin{figure}
    \centering
    \includegraphics[width=\linewidth]{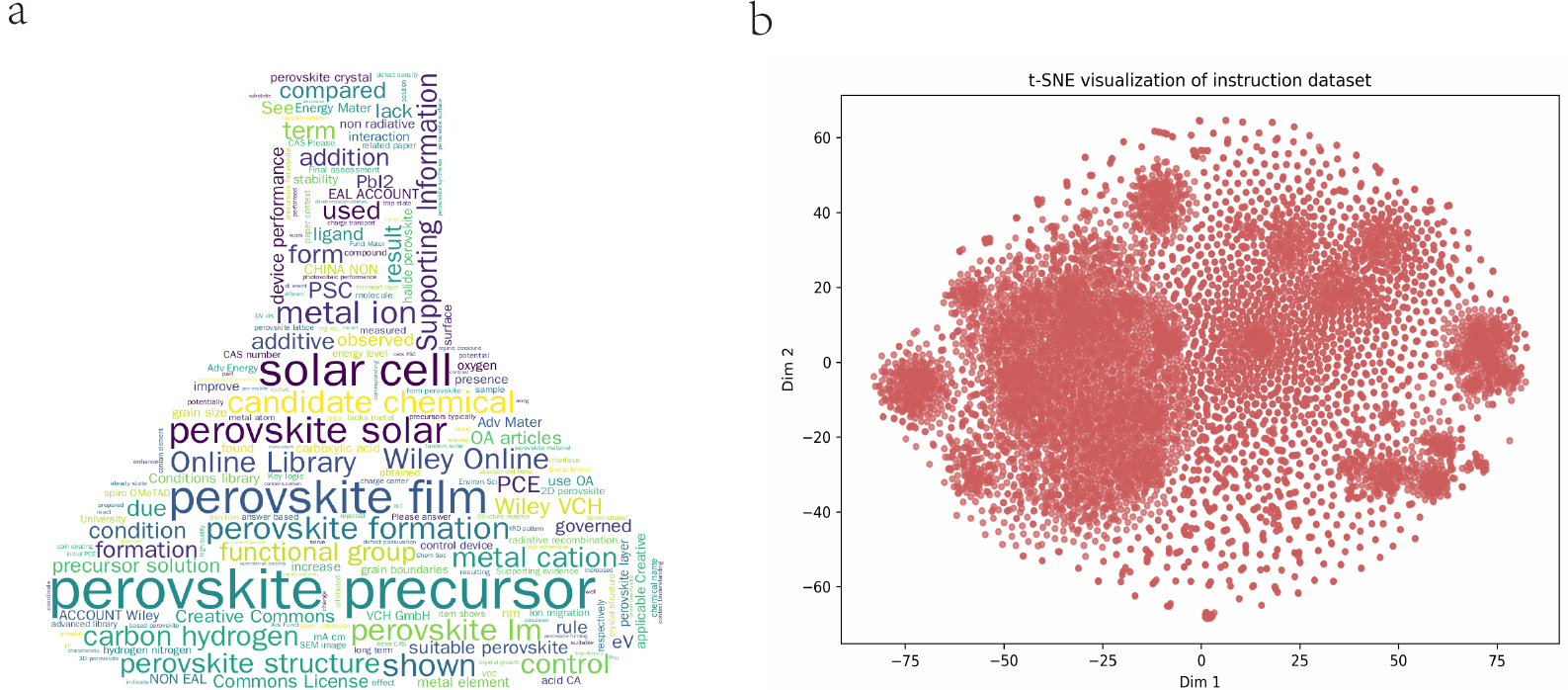}
    \caption{\textbf{The word cloud and t‑SNE visualization of the instruction dataset.} a. The word cloud illustrates the terms that frequently appear in the instruction dataset, including perovskite, precursor and solar, all of which are relevant to the task at the core of this study. b. Two-dimensional t‑SNE map of concatenated input–output TF‑IDF vectors, using perplexity 30, learning rate 200, PCA initialization, and 1,000 iterations. Each point represents a data sample; spatial proximity reflects semantic similarity.}
    \label{fig:fig-5}
\end{figure}

\paragraph{Chain-of-Thought generation}
For each text fragment and drug record, we leverage an LLM to automatically generate paired question–answer instances, each accompanied by detailed CoT reasoning. In the case of the research article data, each segmented passage is used as the input prompt; the model-generated analytical reasoning process is captured as the CoT, while the synthesized summary served as the final answer output. For the drug-related data, the compound names are provided as the primary input, prompting the model to produce a step-by-step analysis of each drug’s properties and mechanisms, followed by an informed assessment of its potential as a perovskite precursor additive as the concluding output.

This systematic approach results in a richly annotated CoT dataset that not only preserves the complexity and depth of scientific reasoning but also actively exercises and reinforces the model’s advanced reasoning capabilities during subsequent fine-tuning and evaluation. By structuring the data in this way, we ensure that the model is trained to deliver interpretable, logically sound, and contextually relevant responses across a diverse range of perovskite research scenarios.

\paragraph{Model fine-tuning}
During the instruction tuning stage, we adopt the QwQ-32B model as our foundation and implement parameter-efficient adaptation using the Low-Rank Adaptation (LoRA) technique across all weight matrices. For optimal performance, LoRA is configured with a rank of 16, an alpha scaling factor of 32, and a dropout rate of 0.1 to mitigate overfitting. Training is conducted using bfloat16 numerical precision, with a per-device batch size of 1, and gradients are accumulated over 8 steps to yield an effective batch size of 8. The learning rate is set to an initial value of 1e-4 and is decayed according to a cosine annealing schedule, incorporating a 5\% warmup phase to stabilize early training dynamics. The model is trained for a total of 10 epochs. To further enhance computational efficiency, we utilize FlashAttention2, which enables rapid and memory-efficient attention calculations. This carefully orchestrated configuration ensures both training stability and the effective integration of domain-specific knowledge into the fine-tuned model.

\section*{Data availability} 
An example of the training set is available from \url{https://huggingface.co/datasets/JH976/Perovskite-R1}. The QwQ-32B model is publicly available through the Qwen organization’s model hub: \url{https://huggingface.co/qQwen/QwQ-32B}, released under Apache License 2.0.

\section*{Code availability} 
The code for splitting the paper to generate the dataset and for calling OpenAI o1 model is available at \url{https://huggingface.co/datasets/JH976/Perovskite-R1}.
The code for fine-tuning the QwQ-32B model is based on the open‑source LLaMA‑Factory framework~\cite{zheng2024llamafactory}. It includes all training scripts, configuration files (e.g., LoRA settings, optimizer, dataset pipelines), and inference utilities. The complete repository is publicly available at: \url{https://github.com/hiyouga/LLaMA-Factory}, released under Apache License 2.0.

\bibliographystyle{unsrt}
\bibliography{references}

\vspace{36pt}
\noindent\textbf{Acknowledgement:}
The work is supported by the National Natural Science Foundation of China (No.62476278, No.11934020). Computational resources have been provided by the Physical Laboratory of High Performance Computing at Renmin University of China.
\\

\noindent\textbf{Author contributions:}  
Z.F.G., C.M., and Z.Y.L. contributed to the ideation and design of the research; X.D.W., Z.R.C., and Z.F.G. performed the research; X.D.W., Z.R.C., Z.F.G., C.M., and Z.Y.L. wrote and edited the paper; all authors contributed to the research discussions. \\

\noindent\textbf{Corresponding authors:} 
Correspondence and requests for materials should be addressed to Ze-Feng Gao (zfgao@ruc.edu.cn), Cheng Mu (cmu@ruc.edu.cn), and  Zhong-Yi Lu (zlu@ruc.edu.cn). \\

\noindent\textbf{Competing interests:}
The authors declare no competing interests.\\

\noindent\textbf{Supplementary information:}
The supplementary information is attached.

\clearpage

\appendix
\renewcommand{\thesection}{\Alph{section}}

\section{An example of prompt}
The following is an example of the prompt utilized in the experiment. The ``Core Design Goal'' section aligns closely with the previously discussed ``Task Definition'' section, ensuring consistency between the design's foundational objectives and the overall task framework. Furthermore, the ``Key Design Principles \& Strategies'' and ``Critical Design Considerations'' sections correspond directly to the ``Objective Criteria'' section, reflecting the essential priorities and strategic approaches established earlier. Finally, the ``Output Requirements'' and ``Additional Filtering Rule'' sections directly map to the ``Expected Output Specification'', providing a clear link between the intended results and the defined output standards. In addition to the aforementioned three sections, the prompt contains several supplementary components that are deemed essential for the comprehensive understanding of the given subject or task.\\

\noindent Core Design Goals:\\
1. Defect passivation.\\
2. Control crystallization kinetics and film morphology.\\
3. Optimize energy-level alignment.\\

\noindent Key Design Principles \& Strategies:\\
1. Lewis acid-base interactions.\\
2. Crystallization modulators.\\
3. Multifunctional additive design.\\

\noindent Critical Design Considerations:\\
1. Solubility in DMF/DMSO.\\
2. Compatibility with perovskite chemistry and subsequent layers.\\
3. Prefer commercially available or natural/food‑grade compounds.\\

\noindent Output Requirements:\\
– Analyze functional groups and mechanistic role.\\
– Evaluate processing compatibility.\\
– Rank 3–5 top candidate additives by expected PV efficiency improvement.\\
– For each: chemical name + CAS number; rationale/mechanism; primary function(s).\\

\noindent Additional Filtering Rule:\\
– **If fewer than 3 candidates truly meet ALL the core goals and critical considerations, do not force the output to contain 3–5 items. Instead, only list those compounds that genuinely satisfy the criteria.**\\

\noindent Here is a list of candidate compounds from our chemical database:\\
- Octa(aminophenyl)-T8-silesquioxane (CAS: 518359-82-5)\\
- 4-Cyanopyridine (CAS: 100-48-1)\\
- Quinuclidine (CAS: 100-76-5)\\
- N-Nitrosopiperidine (CAS: 100-75-4)\\
- potassium phenolate (CAS: 100-67-4)\\
- 3-Pyridinemethanol (CAS: 100-55-0)\\
- N-Benzyl-N,N-dimethyl-1-phenylmethanaminium chloride (CAS: 100-94-7)\\
- 3-Hydroxybenzaldehyde (CAS: 100-83-4)\\
- Butane, 1-(1,1-dimethylethoxy)- (CAS: 1000-63-1)\\
- 1H-Pyrrolo[2,3-b]pyridine-3-carboxaldehyde, 4-bromo- (CAS: 1000340-35-1)\\
- 1H-Pyrrolo[3,2-b]pyridine, 5-bromo- (CAS: 1000341-51-4)\\
- Benzoic acid, 3-amino-5-bromo-2-methyl-, methyl ester (CAS: 1000342-11-9)\\
- 6-Chloro-1H,2H,3H-pyrrolo[3,2-c]pyridin-2-one (CAS: 1000342-80-2)\\
- 1H-Pyrrolo[3,2-c]pyridine, 6-bromo- (CAS: 1000342-71-1)\\
- 1H-Indazole, 6-bromo-5-methyl- (CAS: 1000343-69-0)\\
- 3-Benzofuranacetic acid, 2,3-dihydro-6-hydroxy- (CAS: 1000414-37-8)\\
- 3-Pyridazinecarboxylic acid, 1,6-dihydro-1-methyl-6-oxo- (CAS: 100047-66-3)\\
- 2-Pyridineacetic acid, 5-fluoro- (CAS: 1000515-83-2)\\
- 1H-Indazole, 7-bromo-1-methyl- (CAS: 1000576-59-9)\\
- Benzonitrile, 3-bromo-4-iodo- (CAS: 1000577-94-5)\\

\noindent Please screen and rank these compounds according to the prompt above.\\
**Final Output Instructions:**  \\
– **Output only the selected additives, each enclosed in a single set of braces {}.**\\  
– **Within each brace, include only the chemical name and CAS number.**\\  
– Do not output any extra text or explanation beyond these braces.\\ 
– Ensure that your final selections are fully consistent with your internal reasoning.\\

\section{An example of output}

The following example illustrates the potential efficacy of Perovskite-R1. In light of the example, we have elected to utilize 5-hydroxy-2-methylbenzoic acid, \emph{i.e.} AI-HMBA, as the additive for the ensuing experiment, with the objective of identifying an optimal perovskite additive. The segment between  <think> and </think> represents the chain of thought, which can be regarded as the reasoning process of Perovskite-R1. Meanwhile the segment following </think> is the final output of the input.\\

\noindent<think>

\noindent 1. **Analysis of each compound**:

   - **Dimethylphosphine oxide**:
   
     - **Solubility**: It is likely to be soluble in DMF/DMSO as it has a phosphorus - oxygen bond which can interact with the polar solvents. It has a Lewis basic phosphorus atom that could potentially participate in Lewis acid - base interactions with perovskite precursors. It could act as a ligand or influence the coordination environment during perovskite formation, thus having the potential to control crystallization kinetics.
     
     - **Defect passivation**: The lone pair on the phosphorus atom might be able to interact with defects in the perovskite lattice, potentially passivating them.
     
     - **Energy - level alignment**: It may influence the energy levels through its interaction with the perovskite structure, but more detailed analysis would be needed.
     
   - **5 - Hydroxy - 2 - Methyl - Benzoic Acid**:
   
     - **Solubility**: It has a carboxylic acid group which can hydrogen bond with the solvent molecules in DMF/DMSO, so it is likely to be soluble.
     
     - **Lewis acid - base interactions**: The carboxylic acid group is a Lewis acid, and it could interact with Lewis basic sites in the perovskite precursor, influencing the crystallization process.
     
     - **Defect passivation**: The hydroxyl group and the carboxylic acid group could potentially interact with defects in the perovskite, but the extent and mechanism would require further study.
     
   - **2 - Cyano - 3-(3 - chlorophenylethyl)pyridine**:
     - **Solubility**: The presence of the cyano group and the aromatic rings may make it less soluble in DMF/DMSO compared to some other compounds. The cyano group is a weak Lewis acid, and the pyridine ring has a nitrogen atom with lone pairs that could participate in Lewis acid - base interactions. However, the overall solubility might be a concern.
     
     - **Crystallization control**: The Lewis acid - base interactions could potentially influence the growth of perovskite crystals, but the solubility issue needs to be addressed first.
     
     - **Defect passivation**: The cyano and pyridine groups might interact with defects, but again, solubility is a limiting factor.
     
   - **(1R,4R)-Methyl 4 - formylcyclohexanecarboxylate**:
     - **Solubility**: It has a carbonyl group which can interact with the solvent through hydrogen bonding, so it is likely to be soluble in DMF/DMSO.
     
     - **Lewis acid - base interactions**: The carbonyl group can act as a Lewis acid, and the ester group can participate in some interactions. It could potentially influence the coordination chemistry during perovskite formation.
     
     - **Defect passivation**: The functional groups might interact with defects, but more research on the specific interactions would be needed.
     
   - **Ethyl 3-(3 - nitrophenyl)-3 - oxopropanoate**:
     - **Solubility**: The nitro group and the carbonyl group can interact with the solvent, so it is likely to be soluble in DMF/DMSO.
     
     - **Lewis acid - base interactions**: The carbonyl group can act as a Lewis acid, and the nitro group may influence the electronic properties of the molecule. It could potentially affect the crystallization process.
     
     - **Defect passivation**: The interactions might help passivate defects, but the mechanism would need to be explored further.
     
   - **2,3,5 - Trichlorothiophene**:
   
     - **Solubility**: It has a relatively non - polar structure, so it is less likely to be soluble in DMF/DMSO compared to compounds with more polar functional groups.
     
     - **Lewis acid - base interactions**: The sulfur atom has some potential for weak Lewis acid - base interactions, but overall, due to its solubility limitations, it may not be a good candidate for perovskite precursor additives.
     
     - **Defect passivation**: The interactions are likely to be weak due to the lack of strong polar functional groups.
     
   - **L - Valine, N-(1 - oxopentyl)-N-[[2'-(2H - tetrazol - 5 - yl)[1,1'-biphenyl]-4 - yl]methyl]-**:
   
     - **Solubility**: It has multiple functional groups like the amide, tetrazole, and aromatic rings. The amide and tetrazole groups can interact with the solvent, so it is likely to be soluble in DMF/DMSO.
     
     - **Lewis acid - base interactions**: The tetrazole group can act as a Lewis base, and the amide group can participate in some interactions. It could potentially influence the perovskite crystallization.
     
     - **Defect passivation**: The molecule might interact with defects through its various functional groups.
     
   - **1,2 - Pyrrolidinedicarboxylic acid, 4,4 - diMethyl-, 1-(1,1 - diMethylethyl) ester, (2R)-**:
   
     - **Solubility**: It has ester groups which can interact with the solvent through hydrogen bonding, so it is likely to be soluble in DMF/DMSO.
     
     - **Lewis acid - base interactions**: The ester groups can participate in some interactions, and the molecule could potentially influence the coordination environment during perovskite formation.
     
     - **Defect passivation**: The interactions might help passivate defects, but more study on the specific mechanisms is required.
     
   - **Benzoic acid, 4 - ethoxy-, ethyl ester**:
   
     - **Solubility**: It has an ester group and an ethoxy group which can interact with the solvent, so it is likely to be soluble in DMF/DMSO.
     
     - **Lewis acid - base interactions**: The ester group can participate in some interactions, but it may not have a strong impact on perovskite crystallization compared to some other compounds with more reactive functional groups.
     
     - **Defect passivation**: The interactions are likely to be weak for defect passivation.
     
   - **NU6140**:
   
     - **Solubility**: Without knowing its exact structure in detail, it's difficult to assess solubility. But if it has polar functional groups, it may be soluble in DMF/DMSO.
     
     - **Lewis acid - base interactions**: If it has appropriate functional groups like nitrogen or oxygen with lone pairs, it could participate in Lewis acid - base interactions.
     
     - **Defect passivation**: It might interact with defects, but more information about its structure is needed.
     
   - **Propanoic acid, 3 - butoxy-, butyl ester**:
   
     - **Solubility**: It has an ester group and an alkoxy group which can interact with the solvent, so it is likely to be soluble in DMF/DMSO.
     
     - **Lewis acid - base interactions**: The ester group can participate in some interactions, but it may not have a strong influence on perovskite crystallization.
     
     - **Defect passivation**: The interactions are likely to be weak for defect passivation.
     
   - **tert - butyl N-(4 - bromo - 1,3 - thiazol - 2 - yl)carbamate**:
   
     - **Solubility**: It has a carbamate group which can interact with the solvent, so it is likely to be soluble in DMF/DMSO.
     
     - **Lewis acid - base interactions**: The carbamate group can participate in some interactions, and the bromine atom may influence the electronic properties. It could potentially affect the perovskite crystallization.
     
     - **Defect passivation**: The interactions might help passivate defects, but more research on the specific interactions is needed.
     
   - **11 - Oxa - 2,7,9 - triazatridec - 7 - enoic acid, 3 - carboxy - 8 - [[(1,1 - dimethylethoxy)carbonyl]amino]-12,12 - dimethyl - 10 - oxo-, 1-(9H - fluoren - 9 - ylmethyl) ester, (3S)-**:
   
     - **Solubility**: It has multiple functional groups like carboxylic acid, carbamate, and aromatic rings. It is likely to be soluble in DMF/DMSO due to the presence of polar functional groups.
     
     - **Lewis acid - base interactions**: The carboxylic acid and carbamate groups can participate in Lewis acid - base interactions, potentially influencing the perovskite crystallization.
     
     - **Defect passivation**: The molecule might interact with defects through its various functional groups.
     
   - **2 - Methoxy - 4 - nitrophenol**:
   
     - **Solubility**: It has a hydroxyl group and a nitro group which can interact with the solvent, so it is likely to be soluble in DMF/DMSO.
     
     - **Lewis acid - base interactions**: The hydroxyl group can act as a Lewis base, and the nitro group can influence the electronic properties. It could potentially affect the perovskite crystallization.
     
     - **Defect passivation**: The interactions might help passivate defects, but more study on the specific mechanisms is required.
     
   - **Ethyl N - Boc - 3 - oxopyrrolidine - 2 - carboxylate**:
   
     - **Solubility**: It has an ester group and a carbamate group which can interact with the solvent, so it is likely to be soluble in DMF/DMSO.
     
     - **Lewis acid - base interactions**: The carbamate group can participate in some interactions, and the Boc group may influence the electronic properties. It could potentially affect the perovskite crystallization.
     
     - **Defect passivation**: The interactions might help passivate defects, but more research on the specific interactions is needed.
     
   - **tert - Butyl 3 - cyano - 4 - oxopiperidine - 1 - carboxylate**:
   
     - **Solubility**: It has an ester group and a cyano group which can interact with the solvent, so it is likely to be soluble in DMF/DMSO.
     
     - **Lewis acid - base interactions**: The cyano group can act as a Lewis base, and the ester group can participate in some interactions. It could potentially influence the perovskite crystallization.
     
     - **Defect passivation**: The interactions might help passivate defects, but more study on the specific mechanisms is required.
     
   - **3 - Aminopiperidine - 2,6 - dione hydrochloride**:
   
     - **Solubility**: The presence of the hydrochloride salt may increase its solubility in DMF/DMSO. The amine group can participate in Lewis acid - base interactions.
     
     - **Lewis acid - base interactions**: The amine group can interact with Lewis acidic sites in the perovskite precursor, potentially influencing the crystallization process.
     
     - **Defect passivation**: The interactions might help passivate defects, but more research on the specific interactions is needed.
     
   - **2 - COUMARANONE**:
   
     - **Solubility**: It has an aromatic structure with a carbonyl group. It may have some solubility in DMF/DMSO due to the carbonyl group's ability to interact with the solvent.
     
     - **Lewis acid - base interactions**: The carbonyl group can act as a Lewis acid, and it could potentially interact with Lewis basic sites in the perovskite precursor.
     
     - **Defect passivation**: The interactions might help passivate defects, but more study on the specific mechanisms is required.
     
   - **4 - CHLORO - 6,7 - DIMETHOXY - QUINOLINE - 3 - CARBOXYLIC ACID ETHYL ESTER**:
   
     - **Solubility**: It has an ester group and an aromatic structure with a chlorine atom. It is likely to be soluble in DMF/DMSO due to the ester group's interaction with the solvent.
     
     - **Lewis acid - base interactions**: The carbonyl group in the ester can participate in some interactions, and the chlorine atom may influence the electronic properties. It could potentially affect the perovskite crystallization.
     
     - **Defect passivation**: The interactions might help passivate defects, but more research on the specific interactions is needed.
     
   - **1H - Indole - 3 - butanoic acid, monopotassium salt**:
   
     - **Solubility**: The presence of the potassium salt may increase its solubility in DMF/DMSO. The carboxylic acid group can interact with the solvent.
     
     - **Lewis acid - base interactions**: The carboxylic acid group can act as a Lewis acid, and the indole ring may participate in some interactions. It could potentially influence the perovskite crystallization.
     
     - **Defect passivation**: The interactions might help passivate defects, but more study on the specific interactions is needed.\\

\noindent 2. **Final assessment**:
   - **Top candidates**:
   
     - Dimethylphosphine oxide (CAS: 7211 - 39 - 4): It has a phosphorus - oxygen bond that can interact with perovskite precursors through Lewis acid - base interactions, potentially controlling crystallization kinetics and passivating defects.
     
     - 5 - Hydroxy - 2 - Methyl - Benzoic Acid (CAS: 578 - 22 - 3): The carboxylic acid and hydroxyl groups can participate in Lewis acid - base interactions, influencing crystallization and potentially passivating defects.
     
     - 2 - Methoxy - 4 - nitrophenol (CAS: 3251 - 56 - 7): It has a hydroxyl group and a nitro group that can interact with the solvent and participate in Lewis acid - base interactions, potentially affecting crystallization and defect passivation.\\
     
   - **Other compounds**:
   
     - Compounds like (1R,4R)-Methyl 4 - formylcyclohexanecarboxylate, Ethyl 3-(3 - nitrophenyl)-3 - oxopropanoate, L - Valine, N-(1 - oxopentyl)-N-[[2'-(2H - tetrazol - 5 - yl)[1,1'-biphenyl]-4 - yl]methyl]-, 1,2 - Pyrrolidinedicarboxylic acid, 4,4 - diMethyl-, 1-(1,1 - diMethylethyl) ester, (2R)-, tert - butyl N-(4 - bromo - 1,3 - thiazol - 2 - yl)carbamate, 11 - Oxa - 2,7,9 - triazatridec - 7 - enoic acid, 3 - carboxy - 8 - [[(1,1 - dimethylethoxy)carbonyl]amino]-12,12 - dimethyl - 10 - oxo-, 1-(9H - fluoren - 9 - ylmethyl) ester, (3S)-, Ethyl N - Boc - 3 - oxopyrrolidine - 2 - carboxylate, tert - Butyl 3 - cyano - 4 - oxopiperidine - 1 - carboxylate, 3 - Aminopiperidine - 2,6 - dione hydrochloride, 2 - COUMARANONE, 4 - CHLORO - 6,7 - DIMETHOXY - QUINOLINE - 3 - CARBOXYLIC ACID ETHYL ESTER, 1H - Indole - 3 - butanoic acid, monopotassium salt also show potential but have either solubility concerns or less obvious functional group - based mechanisms for perovskite additive functions compared to the top three candidates.

In conclusion, the top three candidates are Dimethylphosphine oxide, 5 - Hydroxy - 2 - Methyl - Benzoic Acid, and 2 - Methoxy - 4 - nitrophenol, which show promise in meeting the core design goals through their Lewis acid - base interactions and potential for influencing crystallization kinetics and defect passivation. The other compounds also have some potential but require further investigation and optimization of processing conditions to fully assess their suitability as perovskite precursor additives.

\noindent</think>

Dimethylphosphine oxide (CAS: 7211-39-4), 5-Hydroxy-2-Methyl-Benzoic Acid (CAS: 578-22-3), 2-Methoxy-4-nitrophenol (CAS: 3251-56-7) are selected as they are more likely to meet the criteria compared to others, with their functional groups potentially influencing crystallization, defect passivation, and energy-level alignment, and considering solubility and compatibility.

\end{document}